\documentclass{article}

\usepackage{arxiv}

\usepackage[utf8]{inputenc} 
\usepackage[T1]{fontenc}    
\usepackage{hyperref}       
\usepackage{url}            
\usepackage{booktabs}       
\usepackage{amsfonts}       
\usepackage{nicefrac}       
\usepackage{microtype}      
\usepackage{lipsum}
\usepackage{graphicx}
\graphicspath{ {./} }
\usepackage{amsmath}
\usepackage{multirow}
\usepackage{makecell}

\title{An Evaluation of a Visual Question Answering Strategy for Zero-shot Facial Expression Recognition in Still Images}

\author{
Modesto Castrillón-Santana\\
  SIANI - Universidad de Las Palmas de Gran Canaria\\
  Spain\\
  \texttt{modesto.castrillon@ulpgc.es} \\
   \And
Oliverio J. Santana\\
  SIANI - Universidad de Las Palmas de Gran Canaria\\
  Spain\\
  \texttt{oliverio.santana@ulpgc.es} \\
   \And
   David Freire-Obregón\\
  SIANI - Universidad de Las Palmas de Gran Canaria\\
  Spain\\
  \texttt{david.freire@ulpgc.es} \\
   \And
   Daniel Hernández-Sosa \\
  SIANI - Universidad de Las Palmas de Gran Canaria\\
  Spain\\
  \texttt{daniel.hernandez@ulpgc.es} \\
   \And
   Javier Lorenzo-Navarro \\
  SIANI - Universidad de Las Palmas de Gran Canaria\\
  Spain\\
  \texttt{javier.lorenzo@ulpgc.es} \\
}

\newenvironment{acks}
    {\section*{Acknowledments}} 
    {} 

\begin{document}
\maketitle
\begin{abstract}
Facial expression recognition (FER) is a key research area in computer vision and human-computer interaction. Despite recent advances in deep learning, challenges persist, especially in generalizing to new scenarios.
In fact, zero-shot FER significantly reduces the performance of state-of-the-art FER models. To address this problem, the community has recently started to explore the integration of knowledge from Large Language Models for visual tasks. In this work, we evaluate a broad collection of locally executed Visual Language Models (VLMs), avoiding the lack of task-specific knowledge by adopting a Visual Question Answering strategy. 
We compare the proposed pipeline with state-of-the-art FER models, both integrating and excluding VLMs, evaluating well-known FER benchmarks: AffectNet, FERPlus, and RAF-DB.
The results show excellent performance for some VLMs in zero-shot FER scenarios, indicating the need for further exploration to improve FER generalization.
\end{abstract}


\section{Introduction}

Human communication extends beyond spoken words. Intonation, gestures, and facial expressions play a vital role in conveying emotions and enriching human-machine interaction (HMI).
More specifically, facial expressions provide information about the message people are transmitting, their emotional state, and how they perceive us. 
Reading facial signals is fundamental to human empathy and is of great utility for automated systems aiming to enhance user experience in HMI.

Facial Expression Recognition (FER) is the field within computer vision and machine learning focused on the estimation of human emotions based on facial expressions~\cite{EKMAN1975,LI2022}. Even though the problem has already attracted the community with traditional techniques, recent advances, particularly with integrating deep learning techniques, have spurred the performance of FER systems, also avoiding the use of invasive devices. However, there are evident generalization limitations in unrestricted scenarios. In fact, models trained with one dataset do not generalize well to other datasets. This situation, that happens even for state-of-the-art FER models, evidences that cross-dataset FER is still a challenging problem.

In this paper, we focus on zero-shot FER adopting Visual Language Models (VLMs) to solve the task. 
Our contributions include: 1) a comprehensive study of VLMs for zero-shot FER, 2) a VQA-based strategy for creating a state-of-the-art FER model, and 3) demonstrating the need for prompt tuning to leverage task-specific information.

\section{Previous work}

FER literature typically classifies human emotions into six basic emotional categories defined by Paul Ekman~\cite{EKMAN1975}: anger, disgust, fear, happiness, sadness, and surprise, plus the neutral state. Two approaches are mostly adopted depending on if FER is extracted from still images or video sequences~\cite{WangY24-arxiv}. In both scenarios, recent advances in FER are mostly driven by deep learning techniques, as the improvement achieved with handcrafted features has plateaued~\cite{Canal22-is}. However, even the best existing deep learning models do not generalize well when deployed in different unseen scenarios~\cite{Kim23-rtip}. To overcome that limitation, the impressive results reported recently by Large Language Models (LLMs) for different purposes have attracted researchers to transfer LLM knowledge for zero-shot FER. In~\cite{ZhaoZ24-arxiv}, the authors adopted CLIP to extract the image visual embeddings, mapping the initial joint vision-language space with a text instruction-based strategy guided by an independent FER dataset. This action aligns both CLIP and LLM feature spaces with contrastive loss. However, the authors conclude that the FER problem is still challenging. Within the same research group, LLM-based zero-shot FER has also been explored in the dynamic scenario, designing a pre-training strategy integrating vision and text~\cite{Foteinopoulou24-fg}. In this paper, our framework is related in that we also utilize VLMs; however, rather than directly using extracted embeddings, we adopt a Visual Question Answering (VQA) approach and instead leverage the textual answers to map to basic facial expressions.


\section{Proposal}

After the massive impact of LLMs, mainly trained unimodally with vast amounts of text data, the recent integration of multimodal data during training has led to the emergence of VLMs with the ability to handle multi-domain problems without fine-tuning. Those models align visual and textual data, delivering flexible solutions for computer vision tasks.
Among the tasks carried out by VLMs, we adopt VQA, which offers a meeting point for vision-language projects~\cite{Agrawal2015VQAVQ, Barra21-prl}. We may refer to the work by Radford et al.~\cite{Radford21-pmlr}, where the authors fine-tuned a vision encoder with an LLM, aligning visual and linguistic representation spaces. The resulting model is able to perform different VQA tasks related to object classification, action recognition, or optical character recognition, outperforming a supervised ResNet-50 trained on a large domain-specific dataset while reducing computational costs. Their work inspired the development of the WISE Image Search Engine~\cite{Sridhar23-wiki}, which, over a pre-trained OpenCLIP VLM, designed a nearest-neighbor search engine in the resulting feature space, enabling content-based image search with promising results.

\begin{figure*}[thpb]
      \centering
      \includegraphics[width=\linewidth]{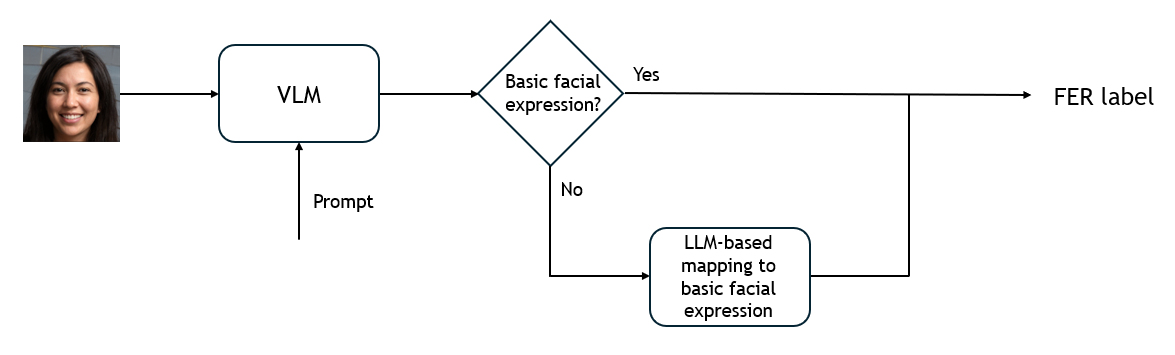}
          \caption{Graphical outline of the sample processing pipeline (image source thispersondoesnotexist.com)}
      \label{fig:pipeline}
\end{figure*}

\subsection{Pipeline}

After the pioneering work by Zhao et al. adopting CLIP in the FER scenario.~\cite{ZhaoZ24-arxiv}, and also inspired by the work by Saha et al.~\cite{Saha24-cvpr}, where the authors made use of prompt-based tuning of VLMs to boost zero-shot categorization in natural domains, our study evaluates different prompts for FER from a VQA perspective. The proposed pipeline estimates facial expressions using a pre-trained VLM, as illustrated in Fig.~\ref{fig:pipeline}. This means that, by adopting a VQA strategy, we explore different questions to be posed to the VLM model for the specific FER scenario. Certainly, below we explore not a single VLM but a wider collection of VLMs for this specific task.

In the outline depicted in Fig.~\ref{fig:pipeline}, a question is referred to as a \emph{prompt}. Below, our prompt exploration covers four different alternatives: 

\begin{description}
    \item - Is the person angry, disgusted, happy, sad, fearful, surprised, or neutral?
    \item - What is the facial expression of the person?
    \item - What facial expression is the person displaying?
    \item - What basic facial expression is the person displaying?
\end{description}

Contrary to a classifier that is trained to provide a restricted set of possible outputs, VQA models are extremely wider in their answers. 
Firstly, the answer is not limited to a single word. In some particular cases, a model such as BLIP-2 OPT~\cite{Junan23-arxiv}, which is well known for its concise answers, is well suited for our problem. However, in many cases, the answer is not limited to a single word and may even include a chain of thought. For that reason, we systematically precede each proposed question with ``\emph{In a single word,}". For instance, the corresponding questions evaluated below in the experimental setup are:

\begin{description}
    \item[\textbf{emoq0}:]~~In a single word, is the person angry, disgusted, happy, sad, fearful, surprised or neutral?
    \item[\textbf{emoq1}:]~~In a single word, what is the facial expression of the person?
    \item[\textbf{emoq2}:]~~In a single word, what facial expression is the person displaying?
    \item[\textbf{emoq3}:]~~In a single word, what basic facial expression is the person displaying?
\end{description}

This simple restriction imposed on the VLM will benefit us by obtaining mostly short answers, reducing the need to analyze extensive responses. However, as we will briefly describe below, it is not completely true that any model will always answerwith a single word even imposing that precondition.




Secondly, given the general-purpose nature of VLMs, their answer will not necessarily be limited to a single word among the six basic expressions (plus neutral). This circumstance happens
 even when suggesting in the prompt the VLM to restrict itself to a set of possible facial expressions (\emph{emoq0}), or refer explicitly to basic facial expressions. According to our experience, in many cases any VLM will not answer one of those basic facial expressions as defined by the FER community. Therefore, for such cases, we have used an independent LLM, specifically ChatGPT\footnote{https://chat.openai.com/chat}, to create a mapping for the answers not included in the limited collection of basic facial expressions. We have thus created a (still incomplete) collection of synonyms in the reported experiments, defining a correspondence of collected answers with the basic ones, i.e., anger, disgust, fear, happiness, neutral, sadness, and surprise. For the experiments described below, using different VLMs and test datasets, the collection of basic facial expression synonyms used is listed in Table~\ref{tab:emotions}.

\begin{table*}[htbp]
\centering
\caption{Set of synonyms adopted to map VLMs answers to the set of basic facial expressions adopted for FER benchmarks annotations. As indicated in the text, we employed ChatGPT to establish the mapping. }
\label{tab:emotions}
\begin{tabular}{@{}>{\bfseries}l p{10cm}@{}}
\toprule
\textbf{Basic emotion} & \textbf{Synonyms} \\
\midrule
Anger & angry, aggressive, aggression, aggravated, derisive, disapproving, evil, frustrated, frustration, mad, pouty, sulky, sulking, stern, yell, yelling. \\
\addlinespace

Disgust & contempt, cringe, disapproval, disdain, disgusted, gagging, grimace, gross, grossed out. \\
\addlinespace

Fear & anxious, anxiety, concern, concerned, covering, fearful, frightened, horror, horrified, intense, nervous, scared, scary, scream, screaming, suspicious, tense, terrified, worry, worried. \\
\addlinespace

Happiness & amused, confident, content, contented, excited, excitement, funny, giggling, goofy, happy, haha, hysterical, joy, joyful, kiss, kissing, kissy, laughter, laughing, laugh, peaceful, satisfied, seductive, silly, singing, slight smile, smiling, smirk, smirking, smug, sticking out their tongue, sticking out tongue, sultry, thumbs up, tongue. \\
\addlinespace

Sadness & agony, anguish, anguished, cry, crying, disappointment, disappointed, discontent, displeased, displeasure, frown, frowning, grief, grim, pain, pained, painful, pout, sad, sorrow, sorrowful, sullen, suffering, unhappy, unsmiling, upset, wistful. \\
\addlinespace

Surprise & baffled, gasp, perplexed, shock, shocked, slightly confused, slightly surprised, surprised. \\
\addlinespace

Neutral & annoyed, bald, bland, blank, bored, boredom, calm, concentrated, concentrating, concentration, contemplation, contemplative, confused, confusion, covered, curious, curiosity, embarrassed, enigmatic, focus, focused, indecipherable, indifference, indifferent, mysterious, mystery, n/a, nosepick, open, peace, pensive, prayer, relaxation, relaxed, sarcastic, sedate, sedated, serious, serene, serenity, shh, shy, skeptical, skepticism, sleeping, sleepy, slightly surprised, speech, speechless, squinting, stupid, sunglasses, tired, thoughtful, thinking, v, yawn, yawning. \\
\bottomrule
\end{tabular}
\end{table*}

In summary, our approach poses a question to the VLM, the reported answer is forced to be short and, if necessary, mapped to one of the basic facial expressions, as depicted in Fig.~\ref{fig:pipeline}.

\section{Experimental evaluation}

After a brief description of the adopted approach, this section summarizes the evaluated questions in seven VLMs and three FER benchmarks considered in the experimental setup.

\subsection{VLMs}

Among the available open VLMs, we have explored possibilities that can be launched locally, ensuring that there is no change in the model being used, as could happen with VLMs in the cloud whose persistence may not be controlled. So far, we have been able to evaluate the following VLMs for the VQA-FER related scenario:

\begin{itemize} \item BLIP-2: A training strategy that integrates off-the-shelf frozen image encoders and language models~\cite{Junan23-arxiv}. The BLIP-2 VQA models have been successfully applied to pedestrian attribute recognition~\cite{Castrillon24-sncs}. In the experiments below, we have evaluated two different models depending on the selection of the frozen language model adopted: 1) 
the Open Pre-trained Transformers (OPT) as language model~\cite{Zhang22-arxiv}, which has been fine-tuned for VQA with the Visual Transformer (ViT) base backbone~\cite{Dosovitskiy21-iclr}, and 2) 
Flan-T5 model~\cite{Chung22-arxiv} as frozen language model, after exhibiting strong few-shot performance on the VQAv2 benchmark. In particular, we have analyzed Flan-T5 XL.

\item Florence-VL: A family of multimodal large language models~\cite{ChenJ24-arxiv} based on the visual representations produced by the vision foundation model Florence-2~\cite{XiaoB23-arxiv}. Florence-VL has proven to be versatile for different purposes after being fine-tuned with a collection of downstream problems. Two available generalist models, \emph{Florence-2-base-ft} and \emph{Florence-2-large-ft}, are adopted below for our VQA-based FER pipeline. The large model is theoretically better suited to achieve higher precision at a higher computational cost.

\item LLAMA 3.2 vision: A collection of multimodal LLMs based on the LLAMA open LLM. The available models have been instruction-tuned for different tasks, including VQA. In the experiments below, we present results for the 11B model via ollama\footnote{https://ollama.com}. 

\item PaliGemma: Follows the trend of PaLI VLMs, this time combining SigLIP vision model and the Gemma LLM to create a sub-3 B VLM maintaining a performance comparable to larger PaLI models~\cite{Beyer24-arxiv}. In the experiments below, we present results with PaliGemma 3b-mix-224 and PaliGemma 3b-mix-448 models. They differ in the resolution adopted for the input image, the second model resize the image to $448 \times 448$ pixels being able to observe higher details at the cost of increasing memory usage.

\end{itemize}

\subsection{Datasets and metrics}

This paper focuses on FER in still images, therefore we have chosen standard benchmarks mostly used in the recent literature:

\begin{itemize}

\item Affect from the InterNet (AffectNet) comprises a large collection of images, more than one million, which were web-crawled from the web with emotion-related keywords~\cite{MOLLAHOSSEINI2019}. The AffectNet7 collection of images was manually annotated for the six basic emotions (anger, disgust, fear, happiness, sadness, and surprise) and neutral faces. Given that the dataset is split into training and validation, we adopted the validation set containing 3999 images for the zero-shot evaluation, which are balanced across the set of facial expressions.

\item FERPlus. After the 2013 Facial Expression Recognition Challenge\footnote{https://www.kaggle.com/datasets/msambare/fer2013}, comprising around 30k $48 \times 48$ cropped facial images labeled with the mentioned seven different facial expressions, but also including contempt and unknown as labels. The presence of some labels is limited in the dataset, e.g. disgust, with other labels being up to 10 times more present. The more recent annotation, known as FERPlus~\cite{Barsoum16-icmi}, included the label provided by 10 human taggers, allowing researchers to have an emotion probability distribution per face. However, the distribution of labels remains unbalanced. The test set, adopted below for evaluation, contains 3573 images.

\item The Real-world Affective Face Database (RAF-DB) is also labeled with the basic emotions and the neutral expression~\cite{LI2019}. RAF-DB was created by gathering images from a social network, which were categorized and reviewed by around 40 human annotators. 
The samples are split into faces showing a single emotion, and those with compound ones. For the evaluation below, we have adopted the aligned test images present in the single emotion subset, containing a total number of 3068 samples.

\end{itemize}

As metrics, the FER literature commonly presents results in terms of Unweighted Average Recall (UAR) and Weighted Average Recall (WAR). UAR corresponds to the average of the respective classes recall, $Recall_i$, without considering the number of instances per class:

\[
\text{UAR} = \frac{1}{C} \sum_{i=1}^{C} \text{Recall}_i \quad \text{where} \quad \text{Recall}_i = \frac{TP_i}{TP_i + FN_i}
\]

$TP_i$ refers to the number of true positives of class $i$ and $FN_i$ stands for the  number of false negatives for class $i$. Instead, WAR averages recall scores across all classes weighted by the number of true instances per class, $N_i$ according to the total number of samples, $N_{total}$.

\[
\text{WAR} = \sum_{i=1}^{C} w_i \cdot \text{Recall}_i \quad \text{where} \quad w_i = \frac{N_i}{N_{\text{total}}}, \quad \text{Recall}_i = \frac{TP_i}{TP_i + FN_i}
\]

WAR is better suited to describe the feasibility of the model given the frequent class imbalance present in FER datasets. 





\begin{table*}
\caption{Weighted (WAR) and Unweighted average recall (UAR) results for zero-shot FER. The best of each dataset and metric is in bold and the second and third best are underlined. For each benchmark, the number of samples evaluated is shown in brackets.}
\label{tab:results}
\begin{center}
\begin{tabular}{c|c|c|c|c|c}
 \multicolumn{2}{c}{ }  & \multicolumn{3}{c}{Test set}  & \\ \cline{3-5}
Model & Training set & AffectNet7 (\# 3499) & FERPlus (\# 3573) & RAF-DB (\# 3068)& Mean \\
\hline
\multirow{3}{*}{ResEmoteNet} & AffectNet7 & - & 0.12/0.08 & 0.15/0.16 & 0.14/0.12 \\
 & FER13  & 0.31/0.31 & - & 0.50/0.34 & 0.41/0.33 \\
 & RAF-DB & 0.27/0.27 & 0.35/0.21 & - & 0.31/0.24 \\ \hline

Exp-CLIP & CAER-S & \underline{0.44}/\underline{0.44} & 0.55/\textbf{0.48} & 0.59/\textbf{0.65} & 0.53/\textbf{0.52} \\ \hline \hline

Model & Question &  AffectNet7 (\# 3499) & FERPlus (\# 3573) & RAF-DB (\# 3068) & Mean \\ \hline
\multirow{4}{*}{BLIP-2 OPT} & emoq0 & 0.27/0.27 & 0.38/0.21 & 0.47/0.31 & 0.37/0.26 \\
 & emoq1 & 0.33/0.33 & 0.57/0.30 & 0.67/0.44 & 0.52/0.36 \\
 & emoq2 & 0.32/0.32 & 0.44/0.26 & 0.62/0.42 & 0.46/0.33 \\
 & emoq3 & 0.28/0.28 & 0.39/0.24 & 0.57/0.35 & 0.41/0.29 \\ \hline

 \multirow{4}{*}{BLIP-2 FLANT5XL} & emoq0 & 0.21/0.21 & 0.38/0.21 & 0.47/0.39 & 0.35/0.27 \\
 & emoq1 & 0.33/0.33 & 0.57/0.30 & 0.59/0.43 & 0.50/0.35 \\
 & emoq2 & 0.34/0.34 & 0.44/0.26 & 0.59/0.43 & 0.46/0.34 \\
 & emoq3 & 0.34/0.34 & 0.39/0.24 & 0.58/0.43 & 0.44/0.34 \\ \hline

\multirow{4}{*}{\makecell{Florence-VL base-ft}} & emoq0 & 0.13/0.13 & 0.35/0.11 & 0.22/0.15 & 0.23/0.13 \\
 & emoq1 & 0.27/0.27 & 0.50/0.18 & 0.52/0.31 & 0.43/0.25 \\
 & emoq2 & 0.26/0.26 & 0.48/0.17 & 0.50/0.31 & 0.41/0.25 \\
 & emoq3 & 0.16/0.16 & 0.36/0.12 & 0.30/0.19 & 0.27/0.16 \\ \hline

\multirow{4}{*}{\makecell{Florence-VL large-ft}} & emoq0 & 0.14/0.14 & 0.35/0.11 & 0.22/0.14 & 0.24/0.13 \\
 & emoq1 & 0.38/0.38 & 0.64/0.30 & 0.62/0.46 & 0.55/0.38 \\
 & emoq2 & 0.36/0.36 & 0.63/0.30 & 0.61/0.44 & 0.53/0.37 \\
 & emoq3 & 0.37/0.37 & 0.62/0.27 & 0.62/0.45 & 0.54/0.36 \\ \hline

\multirow{4}{*}{\makecell{LLAMA 3.2 11B}} & emoq0 & 0.38/0.38 & 0.60/0.36 & 0.68/0.54 & 0.55/0.43 \\
 & emoq1 & 0.41/0.41 & \underline{0.68}/0.38 & \underline{0.73}/0.58 & 0.61/0.46 \\
 & emoq2 & 0.41/0.41 & \underline{0.67}/0.38 & \underline{0.73}/0.58 & 0.60/0.46 \\
 & emoq3 & \underline{0.43}/\underline{0.43} & 0.66/0.39 & \underline{0.73}/\underline{0.60} & 0.61/\underline{0.47} \\ \hline

\multirow{4}{*}{\makecell{PaliGemma 3b-mix-224}} & emoq0 & 0.40/0.40 & \textbf{0.70}/\underline{0.44} & \underline{0.73}/0.56 & \underline{0.61}/\underline{0.47} \\
 & emoq1 & 0.33/0.33 & 0.61/0.40 & 0.69/0.55 & 0.54/0.43 \\
 & emoq2 & 0.36/0.36 & 0.57/0.39 & 0.67/0.56 & 0.53/0.44 \\
 & emoq3 & 0.36/0.36 & 0.58/0.40 & 0.66/0.55 & 0.53/0.44 \\ \hline

\multirow{4}{*}{\makecell{PaliGemma 3b-mix-448}} & emoq0 & \textbf{0.48}/\textbf{0.48} & 0.63/\underline{0.42} & \textbf{0.77}/\underline{0.62} & \textbf{0.63}/\underline{0.51} \\
 & emoq1 & 0.28/0.28 & 0.54/0.37 & 0.64/0.53 & 0.49/0.39 \\
 & emoq2 & 0.29/0.29 & 0.52/0.37 & 0.64/0.53 & 0.48/0.40 \\
 & emoq3 & 0.23/0.23 & 0.50/0.37 & 0.58/0.50 & 0.44/0.37 \\ 

\end{tabular}
\end{center}
\end{table*}

\subsection{Results}

This subsection presents in Table~\ref{tab:results} the zero-shot FER results for different VLM-based approaches following the proposed VQA strategy. For comparison, the Table also comprises results of state-of-the-art approaches without (ResEmoteNet) and with (Exp-CLIP) VLM integration for still images FER.

ResEmoteNet~\cite{RoyA24_spl} currently represents the state-of-the-art for the AffectNet, FERPlus, and RAF-DB benchmarks. It is a model specifically designed for FER tasks, without any integration of LLMs.
The reported ResEmoteNet results are obtained by evaluating separate models, each trained on a different benchmark's training set. Accordingly, Table~\ref{tab:results} includes three distinct results for ResEmoteNet, depending on the dataset used for training.
We intentionally omit results where the test set matches the training benchmark. Instead, we focus exclusively on cross-dataset evaluations, testing on datasets different from those used for training. This approach aims to better assess the model capacity for zero-shot generalization, i.e., its ability to perform on unseen datasets after training on a specific one.
The results highlight ResEmoteNet limited generalization ability across datasets. Specifically, when examining WAR (Weighted Average Recall), the best performance is achieved when the model is trained on FERPlus and tested on RAF-DB.


The next table row presents results for Exp-CLIP~\cite{ZhaoZ24-arxiv}. Exp-CLIP combines visual and textual information, as it makes use of CLIP, integrating FER task-specific information by fine tuning the obtained output with the FER dataset CAERS~\cite{Lee19-cvpr}. Exp-CLIP beats clearly any ResEmoteNet metric in the zero-shot scenario for any benchmark. As already reported by the authors, the Exp-CLIP FER model generalizes better.

\begin{figure*}[htbp]
  \centering
  \begin{minipage}[b]{0.33\textwidth} 
    \centering
    \includegraphics[width=\linewidth]{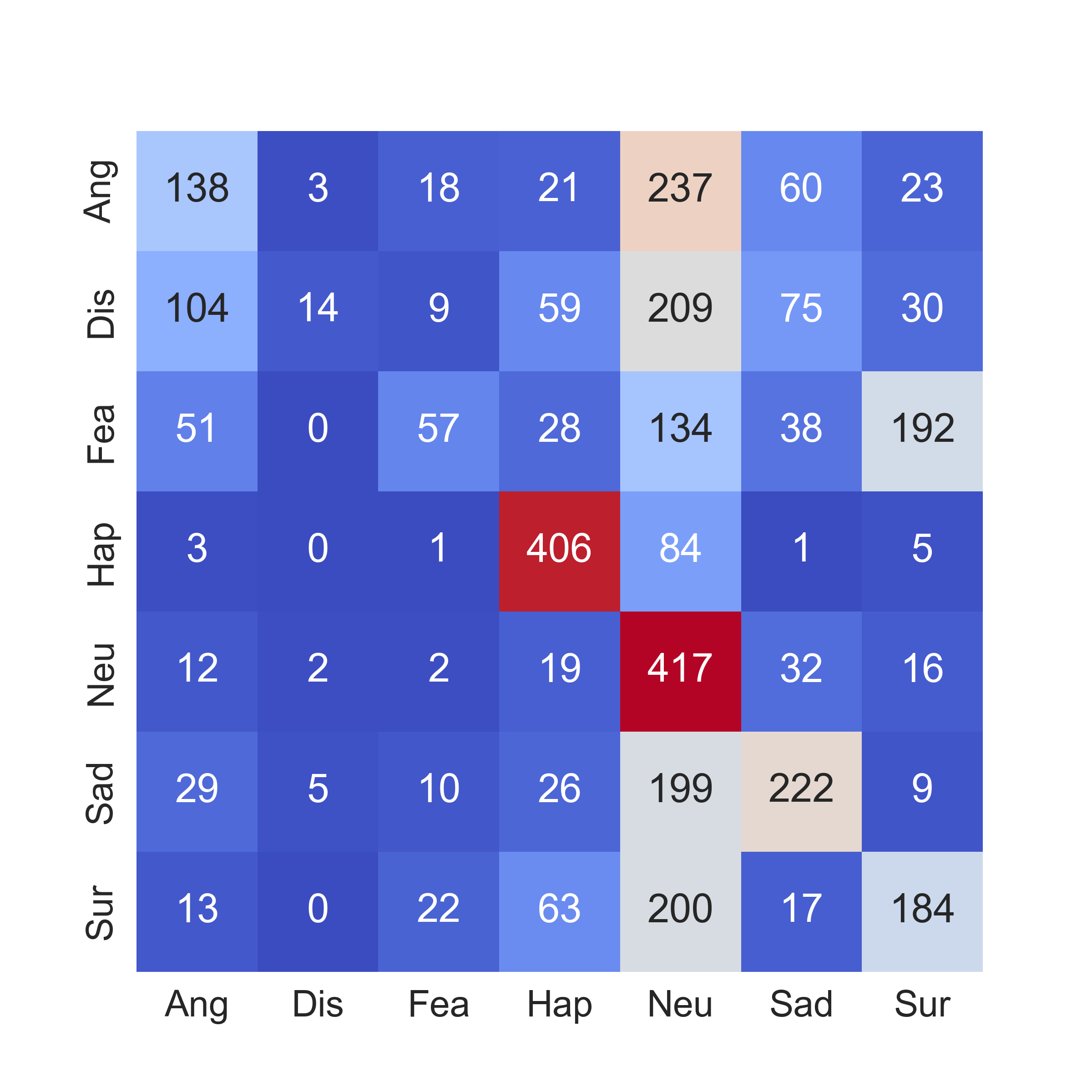} 
    \end{minipage}
  \hfill 
  \begin{minipage}[b]{0.33\textwidth}
    \centering
    \includegraphics[width=\linewidth]{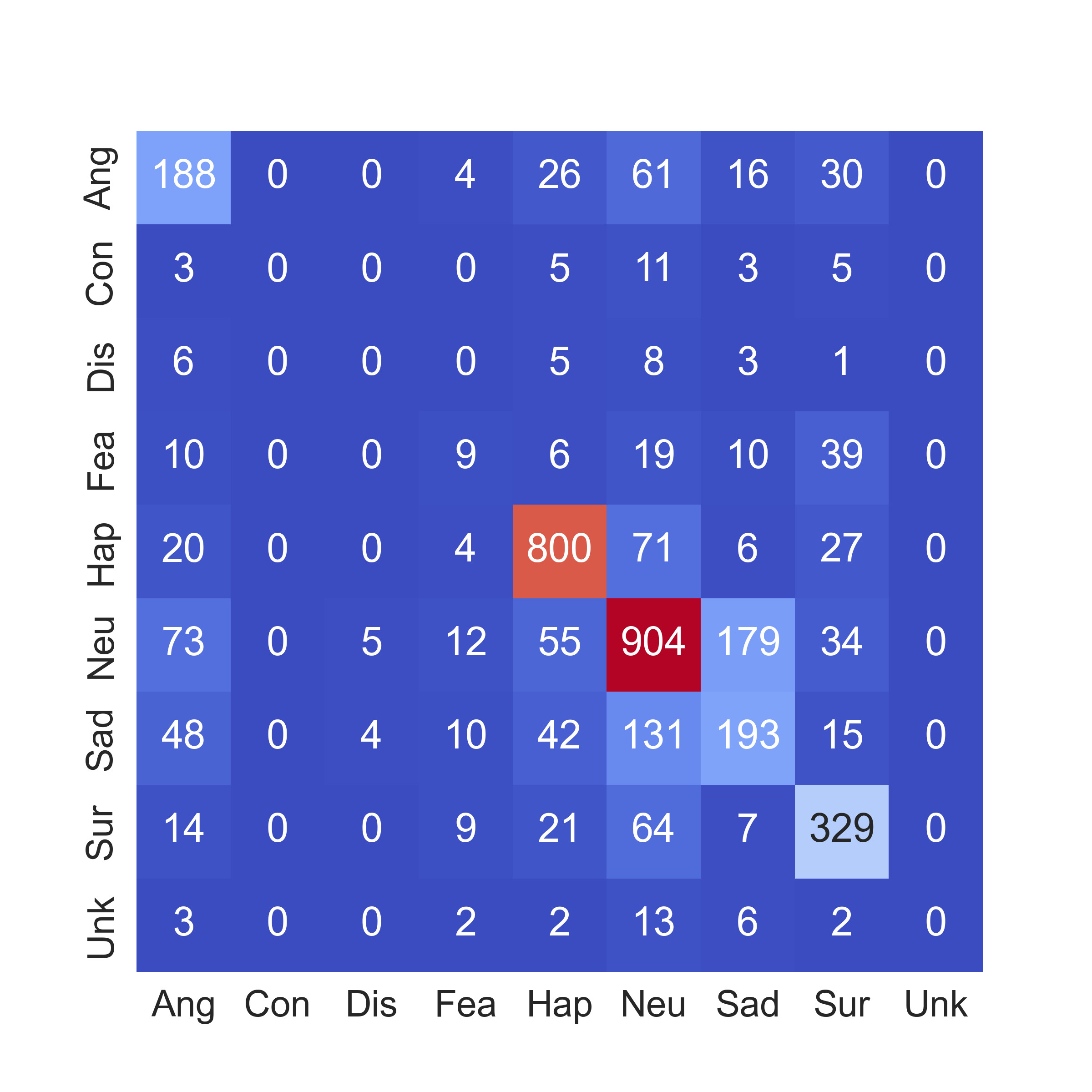}
    \end{minipage}
  \hfill
  \begin{minipage}[b]{0.33\textwidth}
    \centering
    \includegraphics[width=\linewidth]{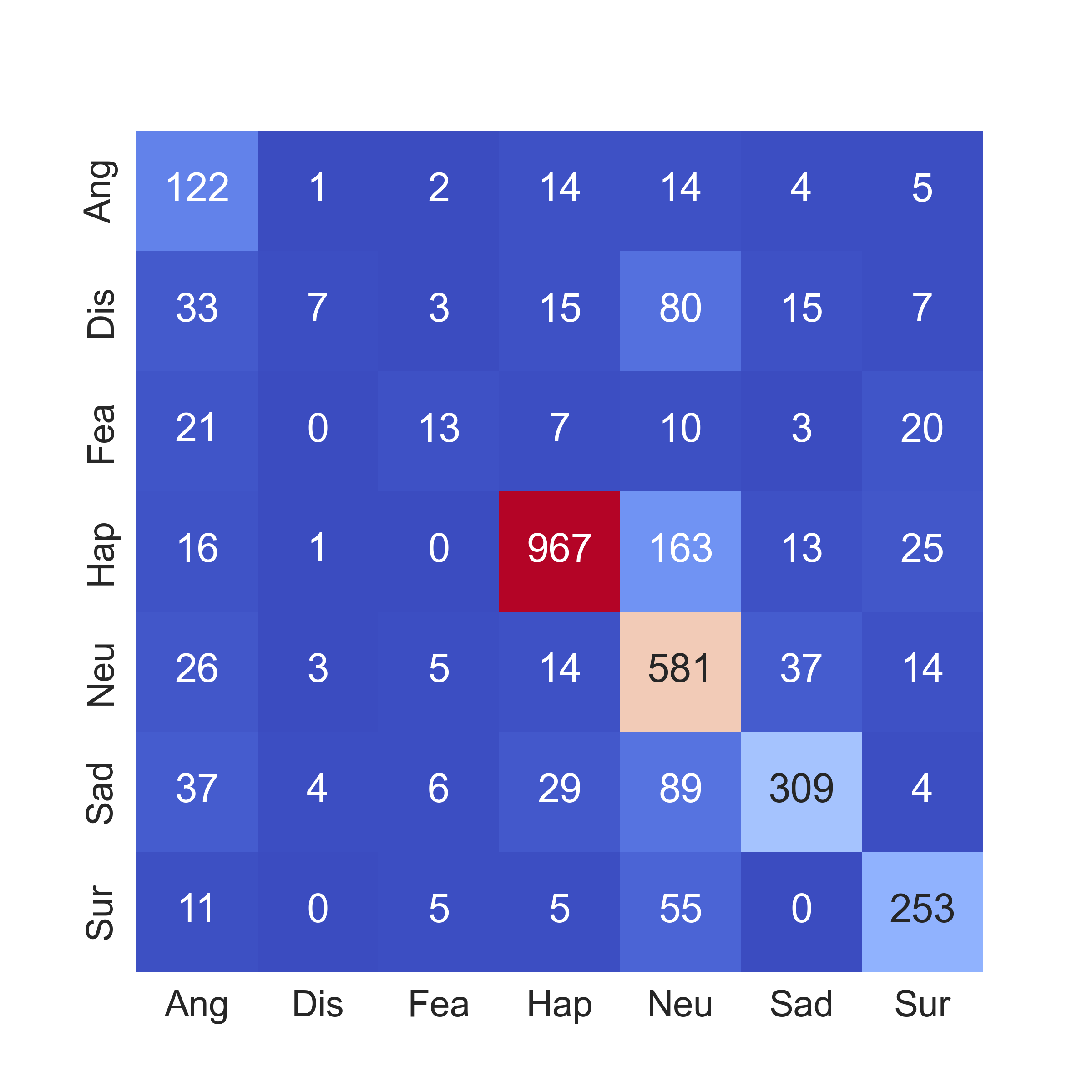}
   \end{minipage}
  \begin{minipage}[b]{0.33\textwidth} 
    \centering
    \includegraphics[width=\linewidth]{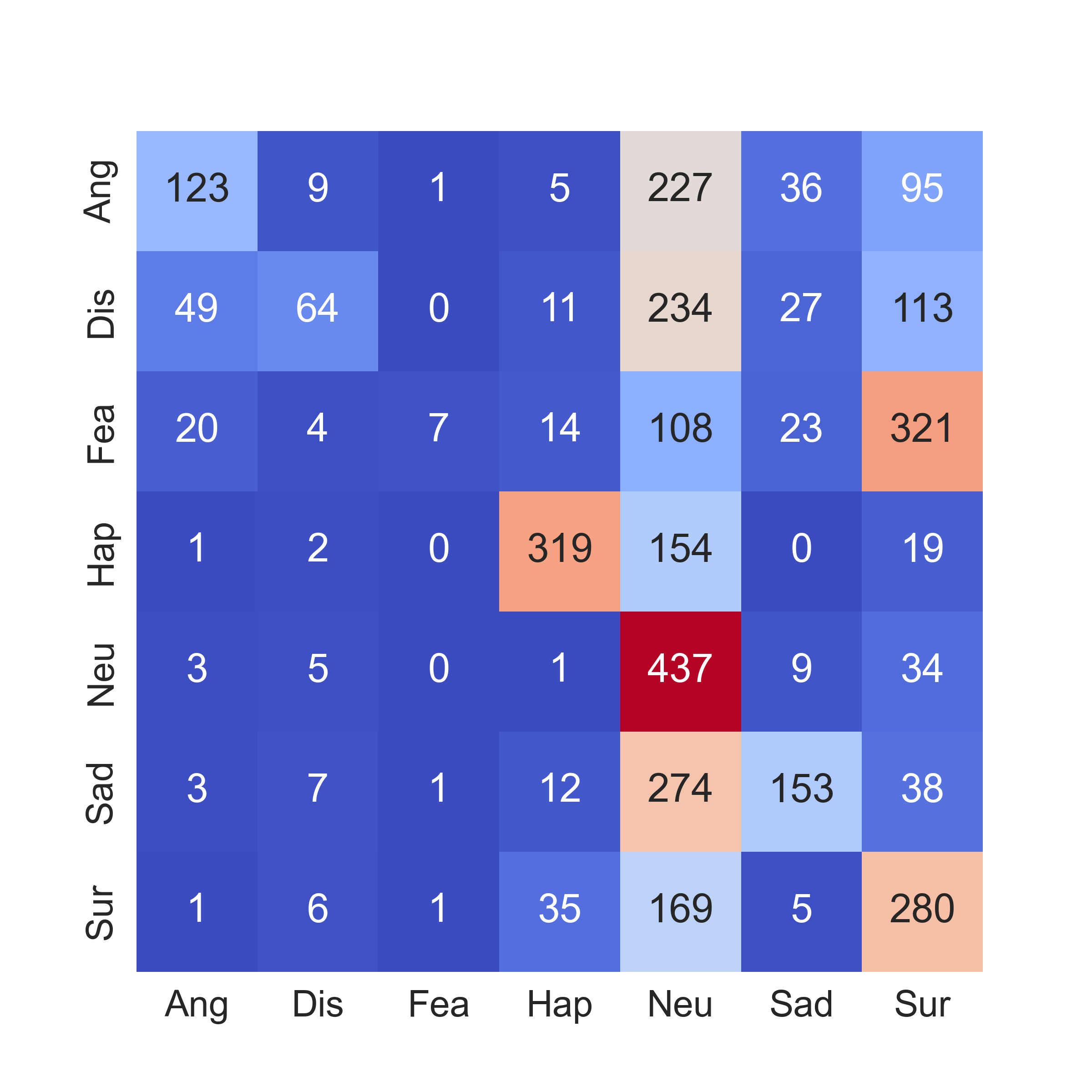} 
    \end{minipage}
  \hfill 
  \begin{minipage}[b]{0.33\textwidth}
    \centering
    \includegraphics[width=\linewidth]{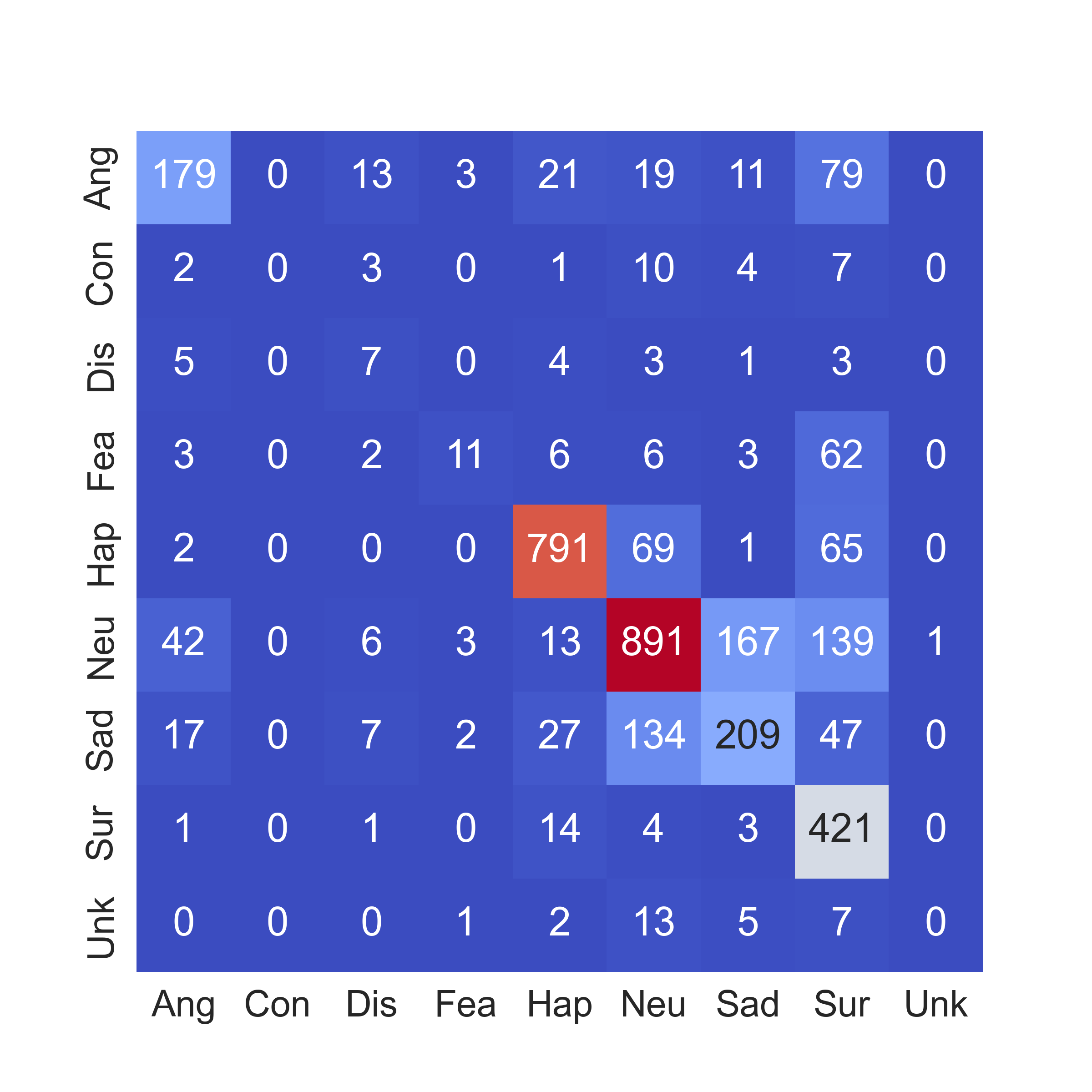}
    \end{minipage}
  \hfill
  \begin{minipage}[b]{0.33\textwidth}
    \centering
    \includegraphics[width=\linewidth]{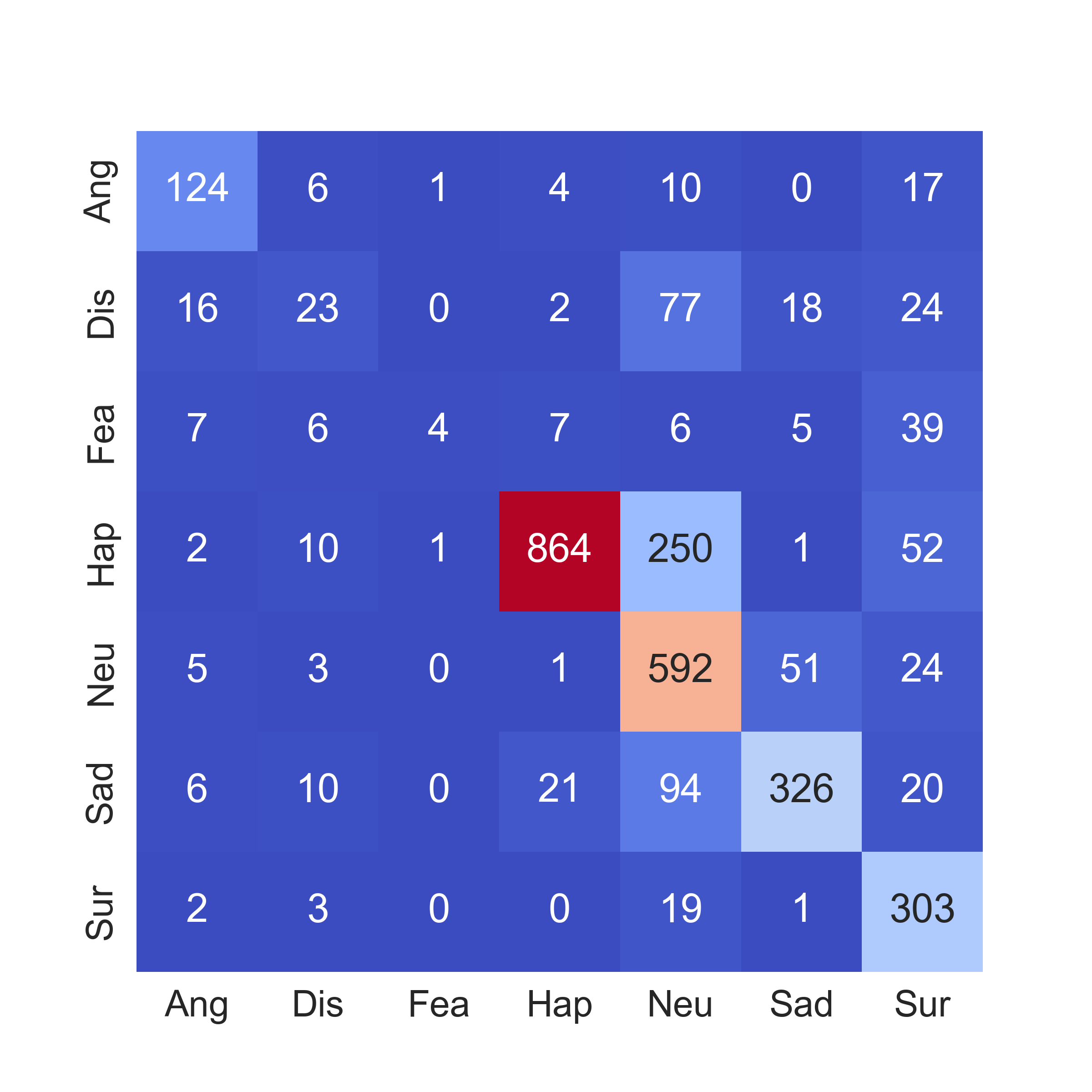}
    \end{minipage}
  \begin{minipage}[b]{0.33\textwidth} 
    \centering
    \includegraphics[width=\linewidth]{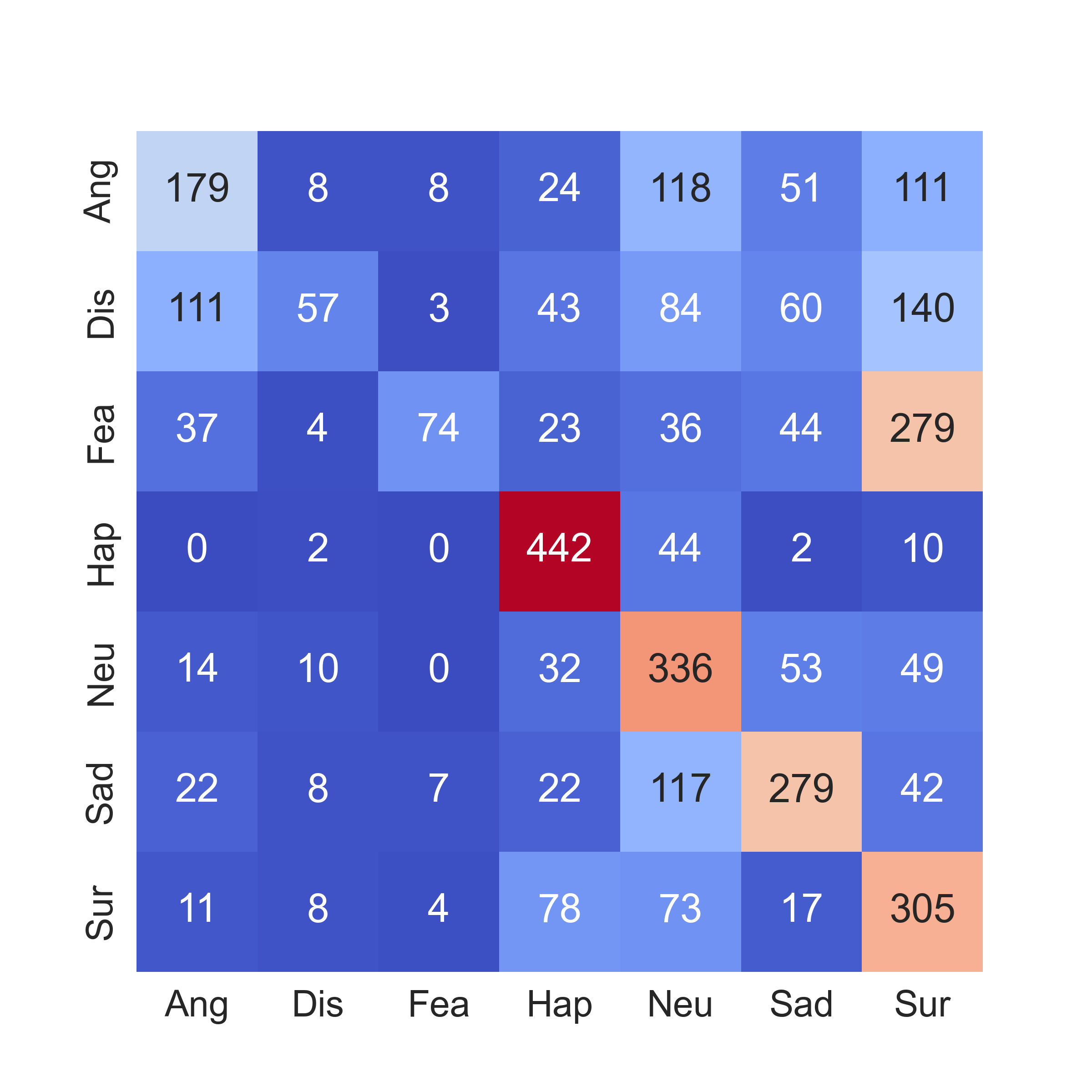} 
    \caption*{(a) AffectNet7} 
  \end{minipage}
  \hfill 
  \begin{minipage}[b]{0.33\textwidth}
    \centering
    \includegraphics[width=\linewidth]{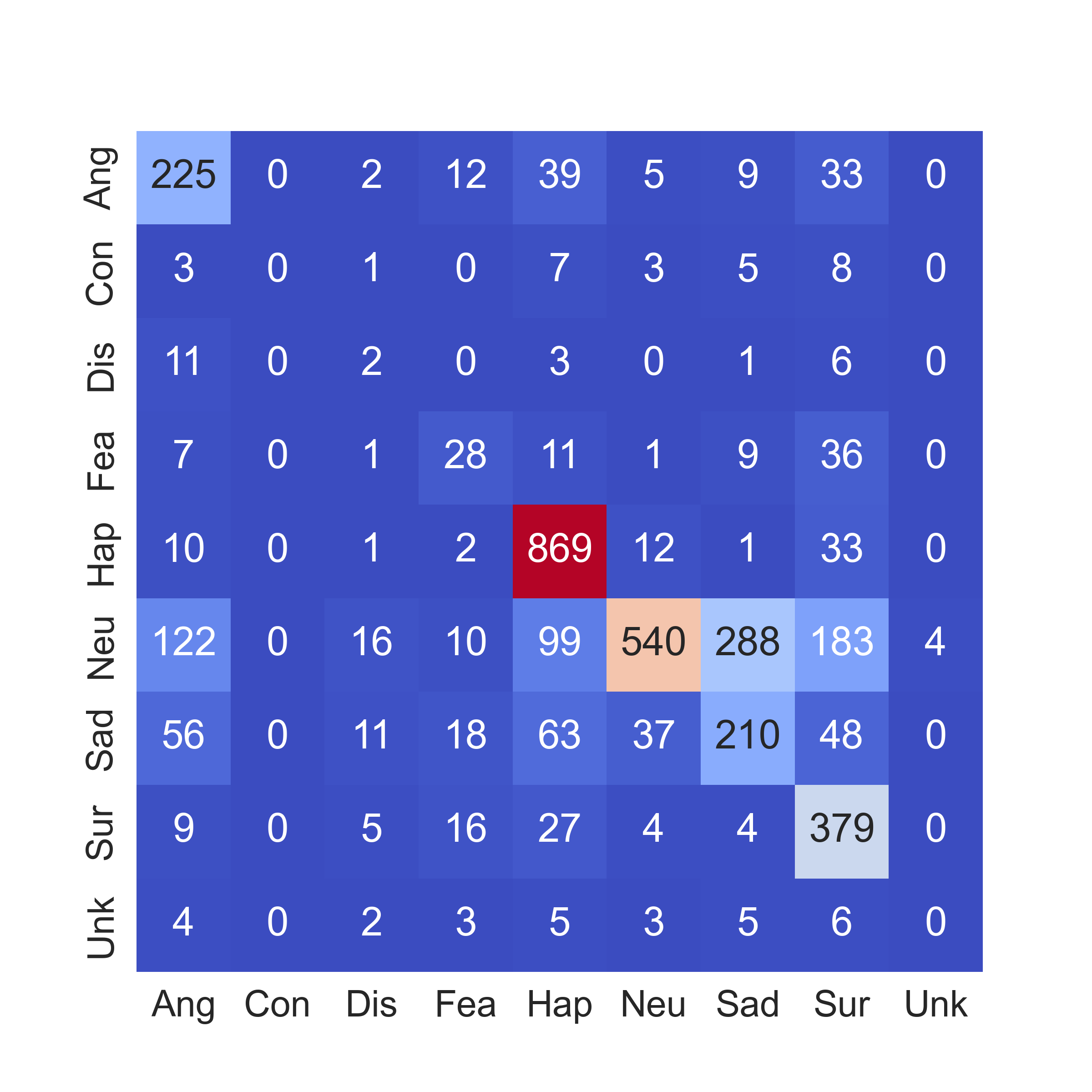}
    \caption*{(b) FERPlus}
  \end{minipage}
  \hfill
  \begin{minipage}[b]{0.33\textwidth}
    \centering
    \includegraphics[width=\linewidth]{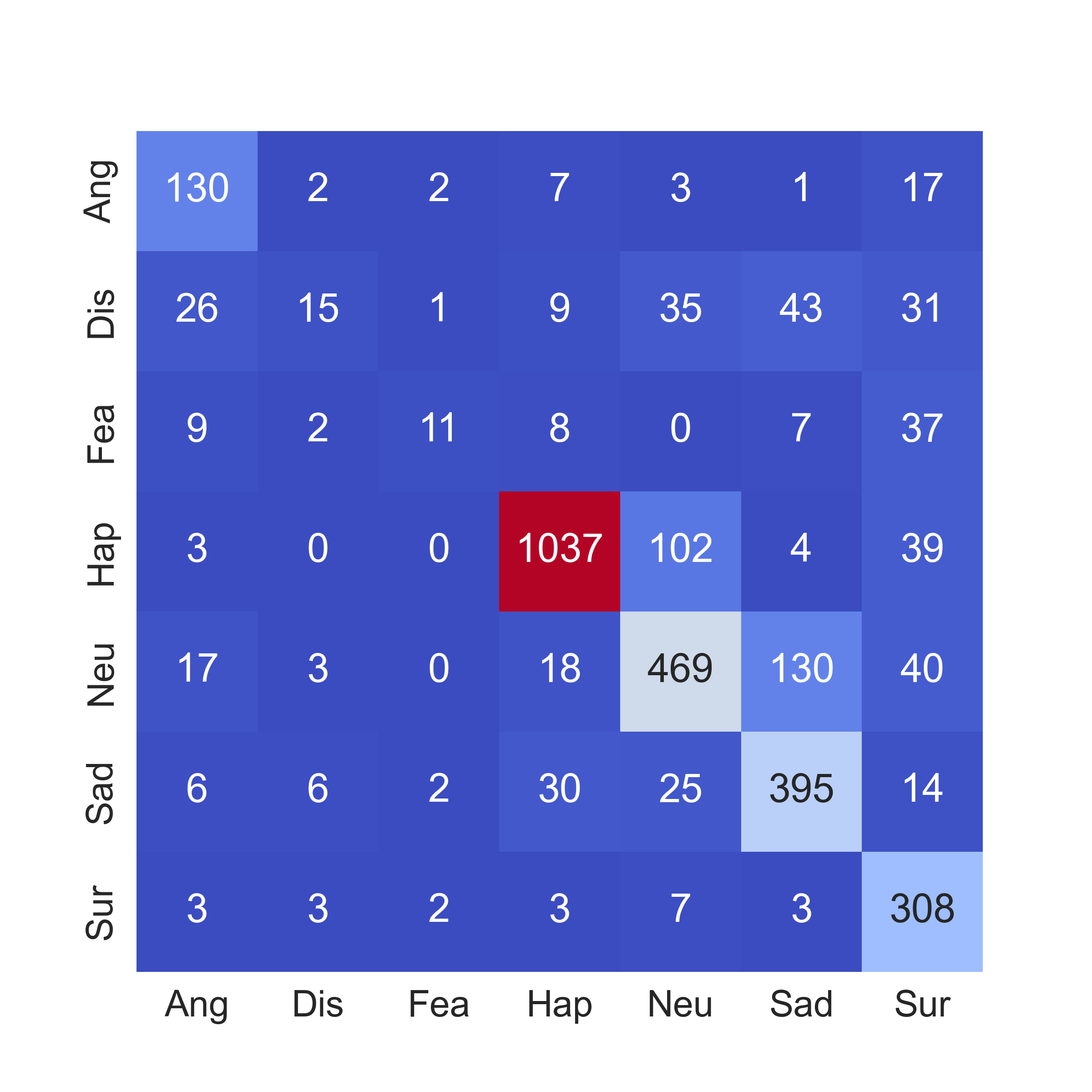}
    \caption*{(c) RAF-DB}
  \end{minipage}
  \caption{Upper row: Confusion matrices for AffectNet7, FERPlus and  RAF-DB with LLAMA 3.2:11B (emoq1). Middle row: Respective confusion matrices for AffectNet7, FERPlus and RAF-DB with PaliGemma 3b-mix-224 (emoq0). Lower row: Respective confusion matrices for AffectNet7, FERPlus and RAF-DB with PaliGemma 3b-mix-448 (emoq0).  
  The labels stand for anger (Ang), disgust (Dis), fear (Fea), happiness (Hap), neutral (Neu) sadness (Sad), surprise (Sur) and unknown (Unk).} 
  \label{fig:ResLLAMA}
\end{figure*}

The second part of Table~\ref{tab:results} summarizes the results of the proposed approach, evaluating up to seven different VLMs without any task-specific training. Instead, task alignment is achieved by selecting targeted, task-related questions, leveraging the VLMs' VQA capabilities to perform the FER task.
The results reveal significant variability in zero-shot FER performance across different architectures. In particular, the choice of prompt plays a crucial role, as certain architectures exhibit strong prompt dependency. For example, Florence-VL large-ft achieves a mean WAR/UAR of 0.24/0.13 with the emoq0 prompt, which increases substantially to 0.54/0.36 with emoq3. In fact, all models showed prompt sensitivity, with emoq1–3 generally yielding better results than emoq0, except for PaliGemma, which performed best with emoq0. This indicates that carefully crafted textual queries are essential for maximizing VLM performance.

Among the evaluated models, PaliGemma 3b-mix-448 achieved the highest average scores, with a WAR of 0.63 and UAR of 0.51 using emoq0. Those results show a WAR ten percentage points higher than Exp-CLIP and 22 larger than the best ResEmoteNet (based on cross-dataset average).
When compared to its lower-resolution counterpart, PaliGemma 3b-mix-224, the results suggest that larger VLMs with higher-resolution inputs (448 pixels) offer advantages in cross-dataset generalization (with the exception of FERPlus), an effect also observed with both Florence-VL variants. However, this improved performance comes at the cost of increased computational demand.
Additionally, the PaliGemma variants exhibited greater performance variance across prompts, in contrast to more stable—but generally lower-performing—models such as BLIP-2 OPT. This suggests a potential accuracy–stability trade-off in model selection.
While slightly below the performance of the PaliGemma models, LLAMA 3.2:11B demonstrated consistent results across emoq1–3, outperforming Exp-CLIP in mean WAR, particularly in RAF-DB and FERPlus. However, its performance was slightly lower on AffectNet7 in terms of accuracy.

Finally, Figure~\ref{fig:ResLLAMA} presents the confusion matrices for the best three VLMs and prompts combination reporting the highest WAR, i.e. LLAMA 3.2:11B with \emph{emoq1} and the PaliGemma variants with \emph{emoq0}. Starting with RAF-DB, models are rather robust for five facial expressions, disgust and fear are rarely well identified. The correct classification of happiness, neutral and sad faces is increased, evidencing the best performance of the largest PaliGemma model. For FERPlus, in addition to disgust and fear, contempt is never identified. VLMs seem not to be aware of it, likely as it is not considered among the basic expressions in the VLMs knowledge. Surprisingly, for this dataset the smallest PaliGemma model is performing better, reflecting the drop in recognizing neutral and surprised faces. AffectNet7 is clearly the most challenging benchmark as no model reaches a WAR of 0.50. Observing the lead reported by the largest PaliGemma model is reflected in the matrices with a less noisy neutral column.


As a final comment, we would like to remark that those results are achieved even when sometimes VLMs models answer is not included in the synonyms table (row \emph{Unk} in the confusion matrix for FERPlus), or the answer is not coherent with a person emotion, e.g. PaliGemma models may answer \emph{Sorry, as a base VLM I am not trained to answer this question} or LLAMA 3.2 may answer \emph{The image is too blurry to determine the person's emotion}.




\section{Conclusions}

FER literature has mostly focused on improving accuracy on existing benchmarks; however, there has been less interest in evaluating zero-shot FER, i.e., testing models with samples from unseen datasets, a challenge that presents great difficulties.
Indeed, state-of-the-art FER models' performance drops remarkably in real-world scenarios.

In this paper, we evaluate different VLMs for zero-shot FER. This is done by avoiding any training, instead prompt engineering under a VQA-based strategy. The use of LLMs is a recent trend in the FER literature, and we are not aware of any broad model evaluation focusing on prompt selection without any modification of the existing pre-trained models. Assuming their biases and limitations, our objective was to evaluate different available VLMs adopting a variety of prompts for the FER problem. 
Our approach shows promising results, achieving the best zero-shot accuracies for well-known FER benchmarks AffectNet7, RAF-DB, and FERPlus. We highlight that this is done with no further training, just prompt fine-tuning.

\begin{acks}
This work is partially funded by project PID2021-122402OB-C22/MICIU/AEI
/10.13039/501100011033 FEDER, UE and by the ACIISI-Gobierno de Canarias and European FEDER funds under project ULPGC Facilities Net and Grant \mbox{EIS 2021 04}. We would like to thank Suave Crew, former Racephotos Sport Photography, for giving us access to the photographs and granting us their usage for research purposes.
\end{acks}

\bibliographystyle{ACM-Reference-Format}
\bibliography{biblio-gias}

\end{document}